\pdfoutput=1
\documentclass[11pt]{article}
\usepackage{acl_24_latex/acl}

\usepackage{times}
\usepackage{latexsym}
\usepackage{multirow}
\usepackage[T1]{fontenc}
\usepackage[utf8]{inputenc}
\usepackage{microtype}
\usepackage{inconsolata}
\usepackage{hyperref}

\usepackage{times}
\usepackage{latexsym}

\usepackage[T1]{fontenc}

\usepackage[utf8]{inputenc}

\usepackage{microtype}

\usepackage{inconsolata}

\usepackage{booktabs}
\usepackage{multirow}
\usepackage{graphicx}

\usepackage{algorithm}
\usepackage{algorithmic}

\usepackage{newfloat}
\usepackage{listings}

\usepackage{subfigure}
\usepackage{amsthm,amsmath,amssymb} 
\usepackage{mathrsfs}
\usepackage{lineno}
\usepackage{xcolor}
\usepackage{color, soul}
\usepackage[inline]{enumitem}
\usepackage{acronym}
\usepackage{enumitem}

\title{MEFT: Memory-Efficient Fine-Tuning through Sparse Adapter}

\author{%
  Jitai Hao\textsuperscript{1} \quad
  Weiwei Sun\textsuperscript{1,2} \quad
  Xin Xin\textsuperscript{1} \quad
  Qi Meng\textsuperscript{3} \\
  \textbf{Zhumin Chen\textsuperscript{1}} \quad
  \textbf{Pengjie Ren\textsuperscript{1}} \quad
  \textbf{Zhaochun Ren\textsuperscript{4}}\thanks{Corresponding author} \\
  \textsuperscript{1}Shandong University, Qingdao, China \\
  \textsuperscript{2}Carnegie Mellon University, Pittsburgh, United States \\
  \textsuperscript{3}Academy of Mathematics and Systems Science, Beijing, China \\
  \textsuperscript{4}Leiden University, Leiden, The Netherlands \\
  \texttt{jitaihao@outlook.com, \{xinxin, chenzhumin, renpengjie\}@sdu.edu.cn} \\
  \texttt{sunnweiwei@gmail.com, meq@amss.ac.cn, z.ren@liacs.leidenuniv.nl}
}

\begin{document}
\maketitle
\begin{abstract}
Parameter-Efficient Fine-tuning (PEFT) facilitates the fine-tuning of Large Language Models (LLMs) under limited resources. However, the fine-tuning performance with PEFT on complex, knowledge-intensive tasks is limited due to the constrained model capacity, which originates from the limited number of additional trainable parameters. To overcome this limitation, we introduce a novel mechanism that fine-tunes LLMs with adapters of larger size yet memory-efficient. This is achieved by leveraging the inherent activation sparsity in the Feed-Forward Networks (FFNs) of LLMs and utilizing the larger capacity of Central Processing Unit (CPU) memory compared to Graphics Processing Unit (GPU). We store and update the parameters of larger adapters on the CPU. Moreover, we employ a Mixture of Experts (MoE)-like architecture to mitigate unnecessary CPU computations and reduce the communication volume between the GPU and CPU. This is particularly beneficial over the limited bandwidth of PCI Express (PCIe). Our method can achieve fine-tuning results comparable to those obtained with larger memory capacities, even when operating under more limited resources such as a 24GB memory single GPU setup, with acceptable loss in training efficiency. Our codes are available at \href{https://github.com/CURRENTF/MEFT}{https://github.com/CURRENTF/MEFT}.

\end{abstract}

\section{Introduction}

\begin{figure}[ht]
\vskip 0.2in
\begin{center}
\centerline{
    \includegraphics[width=\columnwidth]{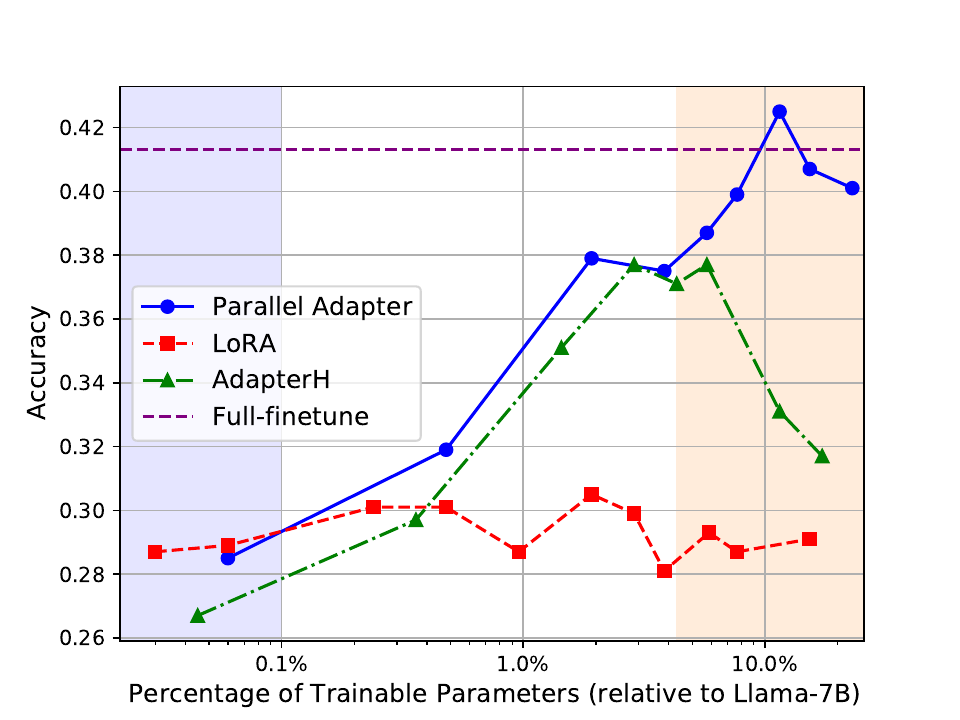}
}

\caption{
Accuracy performance of different PEFT methods on Natural Questions with the rise in the number of trainable parameters. The orange part denotes that the model has reached its fine-tuning limit with a 24GB GPU. The blue part shows performance would be decrease when trainable parameters are limited.
}

\label{progress_performace_nq_v1}
\end{center}
\vskip -0.3in
\end{figure}


As Large Language Models (LLMs) scale up, their fine-tuning becomes increasingly resource-intensive~\citep{touvron2023llama, brown2020language_gpt3}. 
To address this, Parameter Efficient Fine-Tuning (PEFT) methods, which add few trainable modules that only fraction of the LLMs' full parameters (e.g., 0.1\%), have shown promising results on many scenarios, often competitive to full fine-tuning~\citep{adapterH, hu2021lora, li2021prefix, liu2022few_IA3, vucetic2022efficient_far}. 
For instance, LoRA employs matrix decomposition parameters to leverage the low-rank characteristics of LLM parameters, while Parallel Adapter introduces an additional FFN adapter that operates parallel to original FFN layer.


However, the number of new parameters required can vary depending on the task.
For knowledge-intensive tasks, such as QA, it becomes necessary to fine-tune a greater number of parameters to increase the model's capacity for new knowledge memorization~\citep{tirumala2022memorization}.
Figure~\ref{progress_performace_nq_v1} illustrates the performance of LLaMA-7B with Parallel Adapter on NQ datasdet~\citep{nq_data}.
A clear upward trend in answer accuracy is observed as adapter's parameter count increases.
Concretely, for achieving optimal results, approximately 10\% of the parameters are updated, which corresponds to more than 24G of GPU memory usage (the orange part in the figure).

Therefore, reducing the memory cost when the trainable parameters increase is important and has attached research attentions.
Among them, CPU-offload~\cite{ren2021zero_offload_deepspeed} offloads model parameters to CPU memory, which are then returned to the GPU only when computation reaches the corresponding layer. However, CPU-offload transfers all layer parameters, which is inflexible and incurs a high communication overhead.

To address these challenges, we propose MEFT, which introduces \emph{sparse activation} to reduce memory usage. For a given input, sparse activation implies that only a few neurons are activated, as evidenced by various empirical results \cite{liu2023dejavu,geva-etal-2021-transformer} and our pilot experiments detailed at Section \ref{sec2}. 
Inspired by this, MEFT places all trainable parameters in CPU memory to leverage the advantage of the large capacity of CPU memory. 
For a given input, MEFT retrieves a few neurons from CPU memory that are highly relevant to input, and moves them to GPU to complete the computation.
In this way, MEFT significantly reducing GPU memory usage.
Furthermore, to reduces the computational burden on the CPU, we introduce a novel Mixture of Experts (MoE)-based adapter to partition parameters. 
Specifically, given an input, a router is used to activate a subset of the large-scale network and conduct further computation only on these activated subset.
This approach effectively reduces the computational complexity from $O(dNM)$ to $O(dN\sqrt{M})$ to complete the retrieval of relevant parameters.

We conduct experiments on two models (i.e., LLaMA and Mistral) and four datasets (i.e., NQ, SQuAD, ToolBench, and GSM8K).
Experimental results on knowledge-intensive tasks show that our method achieves the best results under resource-restricted conditions. Specifically, MEFT reduces GPU memory usage by 50\% (e.g., reducing from 48G to 24G) while achieving comparable benchmark results.
Our method also outperforms other PEFT methods like Parallel Adapter and LoRA when using the same memory capacity.

In summary, our contributions are as follows:
\begin{enumerate*}[label=(\roman*)]
  \item We propose MEFT, a novel method that utilizes sparse activations and MoE for memory-efficient fine-tuning
  \item MEFT reduces communication overhead by artificially limiting number of activated neurons copied to GPU memory from CPU memory. 
  \item We propose a Key-Experts mechanism to partition a large number of parameters, using a router to allocate inputs to corresponding experts, reducing unnecessary computations and lowering CPU computational burden.
  \item Our experimental results on knowledge-intensive tasks using two popular LLMs demonstrate that our method can achieve the best results under resource-constrained conditions, and the results are comparable to those in relatively resource-rich situations.
\end{enumerate*}

\section{Preliminary}\label{sec2}

PEFT concentrates on fine-tuning LLMs under resource-restricted conditions. To examine performances of different PEFT on knowledge-intensive tasks, we fine-tune LLaMA-7B on Natural Questions with LoRA, AdapterH, and Parallel Adapter. Figure~\ref{progress_performace_nq_v1} shows the results of the experiments. We can see from the left blue part that it is challenging to inject knowledge into the model when the number of trainable parameters is limited (e.g., 0.1\%). Parallel Adapter demonstrates the best scaling results, but for achieving optimal accuracy, approximately 10\% of the parameters are updated. Therefore, we argue that to memorize more knowledge required by downstream tasks, the number of trainable parameters needs to be increased accordingly. Our study is based on Parallel Adapter due to its superior performance. Next, we introduce Parallel Adapter and our motivation for making it more efficient.

\paragraph{Parallel Adapter}
In Transformer-based language models, it has been discovered by researchers that FFNs function as key-value memories \cite{geva-etal-2021-transformer, knowledge_neur}, where each key corresponds to a text pattern, and each value directs the distribution over the output vocabulary. Based on this observation, Parallel Adapter propose to extend the original FFNs by adding a specific knowledge memories tailored for downstream tasks. Specifically, Parallel Adapter places the adapter parallel to the FFNs, and the adapter consists of two linear transformations, $\mathbf{W}_A \in \mathbb{R}^{d\times r}$ and $\mathbf{W}_B \in \mathbb{R}^{r \times d}$, and a nonlinear activation function ReLU.
The computation process of the FFN integrated with the Parallel Adapter can be articulated as follows:
$$
\text{FFN}_{\text{PA}}(\mathbf{h})=f(\mathbf{h} \mathbf{W}_k) \mathbf{W}_v + \text{ReLU}(\mathbf{h} \mathbf{W}_A) \mathbf{W}_B,
$$
where $\mathbf{W}_k \in \mathbb{R}^{d \times n}$ and $\mathbf{W}_v \in \mathbb{R}^{n\times d}$ are weights of the original LLM, $\mathbf{h}$ is the input hidden state of the FFN layer.

\begin{figure}[t]
\begin{center}
\centerline{
\includegraphics[width=0.85\columnwidth]{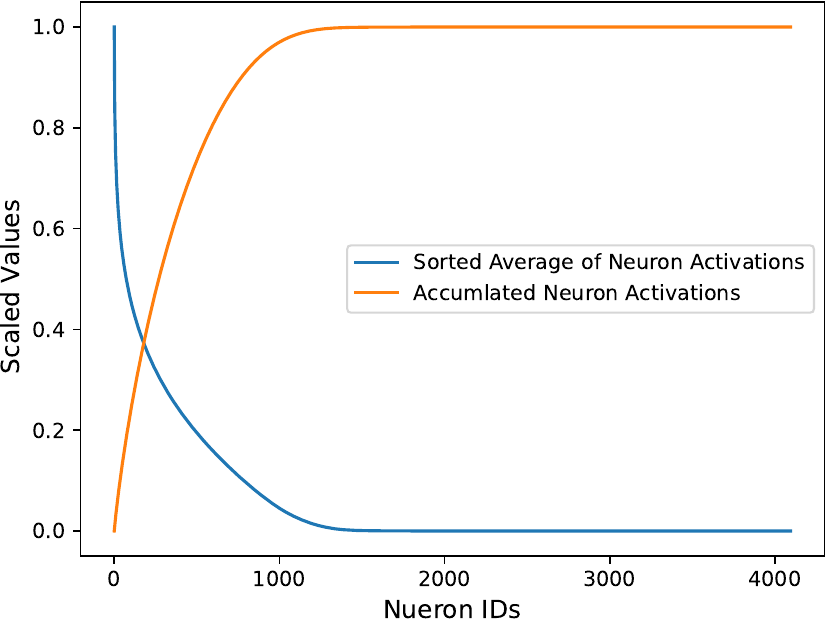}
}
\caption{Sparsity analysis on Parallel Adapter with $4096$ neurons. The neurons are sorted based on activation values. Only a subset of neurons (left part) exhibit high activation value, while majority of neurons are unactivated and not contribute to model's predictions.}
\label{PA_sparse}
\end{center}
\vskip -0.3in
\end{figure}

\paragraph{Sparsity in Parallel Adapter}
As shown in Figure~\ref{progress_performace_nq_v1}, optimizing memory usage presents a challenge when implementing Parallel Adapter with an increased number of parameters for knowledge-intensive tasks. CPU-Offload is a straightforward strategy, where parameters are stored on the CPU and then transferred to the GPU as needed. However, this method transfers all layer parameters, leading to inflexibility and high communication overhead. In this work, we investigate the potential of leveraging activation sparsity in Parallel Adapter. We propose to activate only a select number of neurons for specific inputs, allowing for the transfer of only essential parameters, thereby reducing communication overhead.

We conduct an analysis on sparsity within the Parallel Adapter utilizing the LLaMA-7B model. Specifically, we trained a Parallel Adapter model with a bottleneck size of $4096$ on the Natural Questions dataset. Then we extracted the activation of the adapter's FFNs layer on the test set (which contains $4000$ tokens) and calculated the average activation values. We also computed the cumulative activation values. Note that the neurons are sorted based on their activation value, and the data are normalized to the range [0, 1] for visualization. Figure~\ref{PA_sparse} shows the results. It can be seen that the activation in the adapter is highly sparse, i.e., only a subset of neurons substantially contributes to the model predictions, while the majority of neurons are unactivated. Based on this observation, our work focuses on developing strategies for the effective selection of paramount parameters, and copying only these essential parameters to the GPU during training, thereby reducing the CPU-GPU communication volume and VRAM usage.

\begin{figure*}[ht]
\vskip 0.2in
\begin{center}
\centerline{\includegraphics[width=\textwidth]{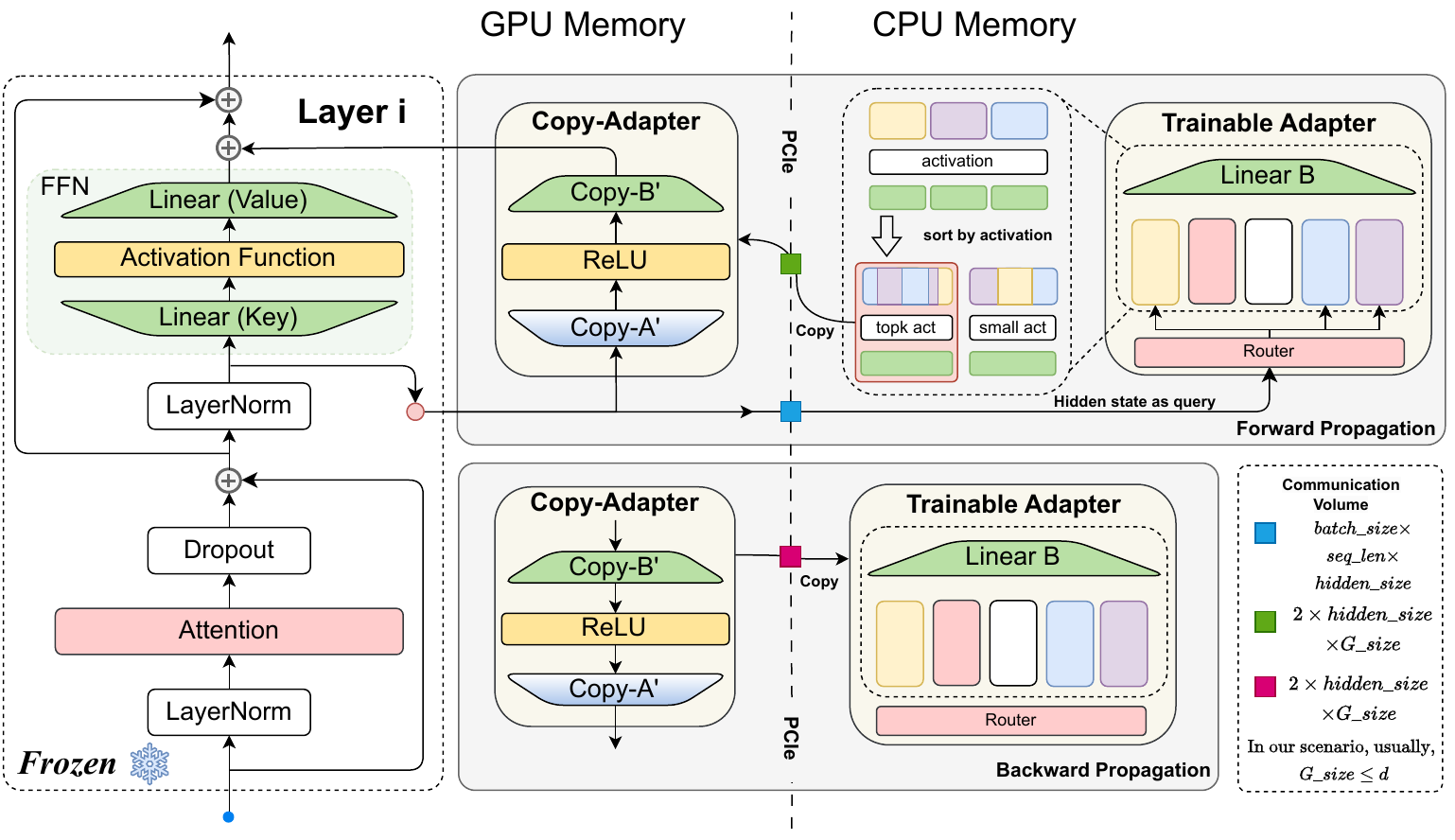}}
\caption{
Overview of our MEFT. The dotted line divides the parameters into two parts, which would be placed on the GPU (left part) and CPU (right part), respectively. Most of the trainable parameters will be allocated to the CPU. During the forward propagation stage, the output of the attention block will be transferred to the CPU to efficiently retrieve neurons highly related to the current context using a MoE-like structure, after which the activated neurons will be transferred to the GPU. During the backward propagation, we transfer the gradients to the CPU and update parameters on the CPU. The above block shown for one Transformer layer is repeated across all the layers.}

\label{overall_pic}
\end{center}
\vskip -0.3in
\end{figure*}

\section{Method}
We propose MEFT, which dynamically loads parameters from CPU memory to GPU to train a larger size Adapter. In Section 3.1, we introduce Sparse Activation to reduce communication overhead, followed by the Key-Experts mechanism in Section 3.2 to reduce the computational complexity of CPU operations. We further analyze the efficiency of our method in Section 3.3.

\subsection{Sparse Activation}
Previous studies have identified contextual sparsity within FFN blocks, attributed to the use of activation functions like ReLU or GeLU~\citep{kurtz2020inducing}. Notably, analyses on various downstream tasks have demonstrated at least 95\% sparsity on FFN neurons, which further leads to sparse gradients. Based on these studies, we explore sparse Adapter training, selectively updating only those neurons that demonstrate higher activation.

Specifically, during the forward computation, $K$ keys in $W_A$ with the highest similarity to $h$ are retrieved and activated for each FFN layer:
\begin{equation}
    S=\text{TopK}(\mathbf{h} \mathbf{W}_A,K),
\end{equation}
where $S$ denotes the indices of the selected keys.
Then we form $\mathbf{W}_A^K$, $\mathbf{W}_B^K$ with related indices on CPU.
\begin{align}
    &\mathbf{W}_A^{K}=\mathbf{W}_A[S] \in \mathbb{R}^{d \times K} \\
    &\mathbf{W}_B^{K}=(\mathbf{W}_B^T[S])^T \in \mathbb{R}^{K \times d}
\end{align}
Here, $\mathbf{W}[\cdot]$ denotes the indexing operation that extracts the corresponding values from matrix $\mathbf{W}$. We aim to extract relevant Keys and Values from $\mathbf{W}_A$ and $\mathbf{W}_B$, respectively.

We then move $\mathbf{W}_A^K$, $\mathbf{W}_B^K$ to GPU as a copy-adapter, then calculate it as widened FFNs.
\begin{equation}
    \text{FFN}_{\text{PA}}(\mathbf{h})=f(\mathbf{h} \mathbf{W}_k) \mathbf{W}_v + \text{ReLU}(\mathbf{h} \mathbf{W}_A^{K}) \mathbf{W}_B^{K} 
\end{equation}

Finally, for backward propagation, only the gradients of these activated neurons are updated, as the non-activated neurons do not contribute to the computation of $\text{FFN}_{\text{PA}}$.

In this way, the majority of the Parallel Adapter's parameters remain stored in the CPU memory, and only these activated neurons need to be copied to GPU memory on-the-fly before each FFN computation.
Given $K \ll r$, where $r$ is the total number of neurons, the activation ratio is typically below $5\%$, allowing us to significantly save GPU memory.

\subsection{Key-Experts Mechanism}
In sparse activation, the $\text{TopK}$ operation that retrieves the most similar weights is executed on the CPU, which may become the bottleneck of computing speed when $r$ is large, given the CPU's lower TFLOPs.
To alleviate this issue, we further propose the Key-Experts mechanism to enhance computation efficiency.

The proposed mechanism is based on the idea of mixture-of-experts, where the weights $\mathbf{W}_A$ and $\mathbf{W}_B$ are divided into $N$ partitions (experts): $\mathbf{W}_{Ai},\mathbf{W}_{Bi},i \in \{1,2,\dots, N\}$, and a router $R(\cdot)$ is employed to route the input to some specific experts.
Specifically, an expert ${E}_i$ is an FFN that consists of $\mathbf{W}_{Ai}$ and $\mathbf{W}_{Bi}$.
For an input token $h$, the router $R(\cdot)$ calculates the score of each expert to be selected:
\begin{equation}
    p_i(\mathbf{h})=\mathbf{W}_g \cdot \mathbf{h}
\end{equation}
Then, the top-$\mathbb{K}$ experts with the highest scores are selected; we use $\tau$ to denote the set of selected indices. The reason why we didn't use softmax to get probabilities is explained in Appendix \ref{sec:negative attempts}.

Then the weights of these selected experts are concatenated to $\mathbf{W}_A^{\prime}$ and $\mathbf{W}_B^{\prime}$:
\begin{align}
\mathbf{W}_A^{\prime} &= \text{Concat}([\mathbf{W}_{Ai}, i \in \tau]) \\
\mathbf{W}_B^{\prime} &= \text{Concat}([\mathbf{W}_{Bi}, i \in \tau])
\end{align}
For this specific token, we can consider $\mathbf{W}_A^{\prime}, \mathbf{W}_B^{\prime}$ as the $\mathbf{W}_A, \mathbf{W}_B$ mentioned in the previous section, but with a smaller size. We then retrieve the top-k key-value pairs to obtain $\mathbf{W}_A^K, \mathbf{W}_B^K$.

Then, we can compute $\text{FFN}_{\text{PA}}(\mathbf{h})$ as shown in Algorithm \ref{ffn_pa_moe}.

\begin{algorithm}[tb]
   \caption{MEFT FFN Layer}
   \label{ffn_pa_moe}
\begin{algorithmic}
   \STATE {\bfseries Input:} hidden state $\mathbf{h}$, original FFN $F$, router $\mathbf{W}_g$, experts of keys $\{\mathbf{W}_A\}$, values $\mathbf{W}_B$, num of experts $\mathbb{K}$, num of key-value pairs $K$.
   \STATE $S \leftarrow \emptyset$ {\quad\small \emph{// initialize S as an empty indices of kv pairs}}
   \FOR{$t$ in $\mathbf{h}$}
       \STATE $p \leftarrow \mathbf{W}_g(t)$ {\quad\small \emph{// gating probabilities}}
       \STATE $\tau \leftarrow \text{TopK}(p, \mathbb{K})$ {\quad\small  \emph{// indices of top$\mathbb{K}$ experts}}
       
       \STATE $\mathbf{a} \leftarrow []$ {\quad\small  \emph{// initialize a as an empty array}}
       \FOR{$i$ in $\tau$}
           \STATE $\mathbf{a}_i \leftarrow \text{ReLU}(\mathbf{h} \mathbf{W}_{Ai})$ {\quad\small \emph{// sparse activation}}
           \STATE $\mathbf{a} \leftarrow \text{Concat}(\mathbf{a}, \mathbf{a}_i)$ {\quad\small  \emph{// gather activation}}
       \ENDFOR
       \STATE $I \leftarrow \text{TopK}(\mathbf{a}, K)$
       \STATE $S \leftarrow S \cup I$
    \ENDFOR

   \STATE $ \mathbf{W}_A^K, \mathbf{W}_B^K \leftarrow (\mathbf{W}_A^T[S])^T, \mathbf{W}_B[S]$
   \STATE // move $\mathbf{W}_A^K, \mathbf{W}_B^K$ to GPU
   
    \textbf{return} $F(h) + \text{ReLU}(h\mathbf{W}_A^K)\mathbf{W}_B^K $
\end{algorithmic}
\end{algorithm}

\subsection{Efficiency Analysis}
While our method could reduce the GPU memory usage as only a fraction of activated neurons are placed on the GPU, the CPU-GPU communication and the CPU computation may cause GPU waiting.
In this section, we analyze the communication volume and the computational complexity of our method.

\paragraph{Communication Volume}
As shown in Figure \ref{overall_pic}, the parameter communication volume between the CPU and GPU consists of two parts: forward and backward propagation. In the following, we use $B$ to denote the batch size, $l$ to denote the length of sequences in the batch, and $\beta$ as a sparsity factor related to $l$.
\begin{itemize}
    \item \textbf{Forward.} For each layer, the hidden state $h$ needs to be moved from GPU to CPU, which causes $B \times l \times d$ communication. Then, after the parameter selection, the activated parameters, with a size of $2 \times d \times \beta \times K$, are moved from CPU to GPU. Here $\beta$ is a sparsity factor related to $l$\footnote{In our experiments, we found that the tokens in a sentence tend to activate similar experts, leading to a smaller $\beta < 0.49$ (experimental results when $B=2,l=256$).}:$\beta = \frac{||\text{Unique}(\text{TopK}(\mathbf{h} \mathbf{W}_A, K))||}{K}$
    
    \item \textbf{Backward.} For each layer, the gradients of the activated parameters calculated on the GPU are moved to the CPU to update the CPU counterpart. Thus the communication volume is the size of the activated parameters, i.e., $2 \times d \times \beta \times K$.
\end{itemize}
Therefore, the total communication overhead of model training is:
\begin{equation}
\text{cost} = nl \times ( 2 \times d \times \beta \times K + B \times l \times d).
\end{equation}

\paragraph{Computational Complexity}
The additional computation on the CPU includes computation on the router and the TopK operation.
Based on the proposed key-experts mechanism, the complexity on the CPU is $MNd + \mathbb{K}\frac{rMd}{N}$.
Therefore, when $N$ approaches $\sqrt{r}$, we can achieve optimal computational complexity $O(dM\sqrt{r})$. This significant reduction in computation makes it well-suited for CPU execution.

\paragraph{Empirical Results}

Our empirical results show that when using LLaMA-7B as the base model, with key-value pairs of size $6144$, a batch size of 2, and a sequence length of 256, the total bidirectional communication volume per batch is about $0.56\mathcal{M}$ per iteration, where $\mathcal{M}$ represents the total number of trainable parameters added to Llama-7B.
In contrast, when using deepspeed-offload \cite{ren2021zero_offload_deepspeed} with the same trainable size, the Parallel Adapter requires a communication volume of $2\mathcal{M}$ per iteration.
Therefore, our method achieves a $3.57\times$ reduction in GPU-CPU communication.

Regarding training efficiency, empirical results show that our method achieves at least 63\% speed compared to the baseline that excludes the time cost of additional communication and CPU computation.
Moreover, there remains the potential to further improve efficiency through engineering methods as detailed in Appendix \ref{sec:implementation_detail}.


\begin{table*}[htbp]
\centering\setlength\tabcolsep{5pt}
\begin{tabular}{clcccccc}
\toprule

\textbf{Model} & \textbf{Method} & \textbf{VRAM}  &\textbf{Param.} & \textbf{NQ} & \textbf{SQuAD} & \textbf{Tool}& \textbf{GSM8k}  \\ 

\midrule

\multirow{9}{*}{\textbf{LLaMA-7B}} & Base Model & - & -& 0.164 & 0.120 &  - & 0.110   \\

\cmidrule(lr){2-8}& LoRA         & 24G &2\%& 0.305 & 0.162 & 0.102 & 0.367\\
 & AdapterH & 24G &2\%& 0.377 & 0.190 & 0.645& 0.525  \\
 & Parallel Adapter& 24G &2\%& 0.387 & 0.236 &  0.636 &   \textbf{0.563}\\
 
 
 & \textbf{MEFT (Ours)} & 24G &10\%& \textbf{0.413}& \textbf{0.290}&  \textbf{0.646}&   0.515\\
\cmidrule(lr){2-8}& LoRA         & 48G &10\%& 0.293 & 0.126&  0.152&   0.311\\
 & AdapterH & 48G &10\%& 0.391 &           0.230 &  \text{0.662}&   0.506\\
 & Parallel Adapter& 48G &10\%& \text{0.425} & \text{0.295}&  \text{0.639}&   0.502\\
\cmidrule(lr){2-8} & Full Fine-tune& -& 100\%& 0.413& 0.313& 0.796&0.602\\

\midrule

\multirow{5}{*}{\textbf{Mistral-7B}}
&  Base Model &  - & -&  0.373&  0.173&  -&   0.522\\ 
\cmidrule(lr){2-8}& LoRA&  24G & 2\%& 0.381& 0.170& 0.345&\textbf{0.715}\\
 & AdapterH&  24G & 2\%& \text{0.415}& 0.156& 0.582&0.707\\
 & Parallel Adapter& 24G & 2\%& 0.401& \text{0.198}& \text{0.762}&0.700\\
 & \textbf{MEFT (Ours)} & 24G & 10\%& \textbf{0.427}& \textbf{0.224}& \textbf{0.772}&\text{0.709}\\

 \bottomrule
\end{tabular}

\caption{
Performance of different methods on downstream tasks. 
VRAM represents the GPU memory required for training. 
Param. shows the percentage of trainable parameters of the model.
"Base Model" indicates the zero-shot performance of the original model on the tasks. The base model can't answer APIs without fine-tuning on ToolBench.
\textbf{Bold} signifies the best result under 24GB VRAM.
}
\label{tab:performance_comparsion}
\vspace{-3mm}
\end{table*}

\vspace{-3mm}
\section{Experimental Setup}

\subsection{Datasets}
We consider four datasets for experiments: Natural Question, SQuAD, ToolBench, and GSM8K:
\begin{itemize}
    \item \textbf{Natural Questions}~\cite{nq_data} is an open-domain question answering dataset. We are considering a close-book setting, i.e., the models are trained to answer questions without background passages~\citep{wang2021can_close_book_qa}.
    
    \item \textbf{SQuAD}~\cite{rajpurkar-etal-2016-squad} is a reading comprehension dataset, consisting of questions posed by crowdworkers on a set of Wikipedia articles. The same close-book setting like NQ is used in SQuAD too.

    \item \textbf{ToolBench}~\cite{qin2023toolbench} is a tool learning dataset, which includes 16K APIs and 432K related queries. The task aims to predict the name and info of APIs that solve the query. 

    \item \textbf{GSM8K}~\cite{cobbe2021training_gsm8k} is a dataset of 8.5K high-quality diverse grade school math word problems. 
    We use MetaMathQA \cite{yu2023metamathqa} as training data for GSM8K.
\end{itemize}

\subsection{Metrics}
For NQ and SQuAD, we use Exact Match (EM) to measure whether the standard answer appears in the model's output.
For ToolBench, we use Intersection over Union (IoU) as the metric, taking the intersection and union of the APIs in the model's answer set $\mathbb{Y}$ and the ground truth set $\mathbb{G}$, and then calculate as: $\text{IoU}=\frac{\mathbb{Y} \cap \mathbb{G}}{\mathbb{Y} \cup \mathbb{G}}$.
For GSM8k, we judge whether the final answer given by the model (matched by regular expression) is correct.

\subsection{Implementation Details}
We utilized the publicly pre-trained LLaMA-7B \cite{touvron2023llama} and Mistral-7B \cite{jiang2023mistral} as the base model. 
All experiments are implemented using PyTorch and the Hugging Face Trainer, on machines with RTX-3090 or A40 GPUs. During fine-tuning, we employ the Adam optimizer for all experiments. For LLaMA-7B, we trained models with the learning rate progressively increasing over the initial 2\% of the steps to $1e^{-4}$, and then decaying linearly to $0$. For Mistral-7B, we trained models with a peak learning rate of $1e^{-6}$ to ensure numerical stability. For tasks NQ, ToolBench, and GSM8k (MetaMathQA), we trained models with an accumulated batch size of 64. We set the number of activated key-value pairs $K=64$ for all tasks, and the number of activated experts $\mathbb{K}=4\sqrt{r}$.
The training epochs for the tasks NQ, SQuAD, Tool, and MetaMathQA (GSM8k) are set to 4, 8, 1, and 1, respectively. Each experiment is conducted on an RTX 3090, taking approximately 12 GPU hours, and the results are reported from a single run.
For task SQuAD, we used a higher batch size of 256 for better performance across all methods. For baselines with LLaMA-7B, we tested their performance with the rank/bottleneck size set to \{32, 64, 128, 256, 512, 1024, 2048\} to achieve optimal performance. It's important to note that a rank larger than 512 and a bottleneck size larger than 256 require 48 GB of GPU memory. For baselines with Mistral-7B, we tested the rank/bottleneck size set to 32, 64, 128, and 256, but only within a 24 GB setting. For most experiments and hyperparameter studies, we report our results from MEFT on Natural Questions.

\begin{figure*} 
\centering 
\begin{minipage}{.45\textwidth} 
\centering 
\includegraphics[width=\columnwidth]{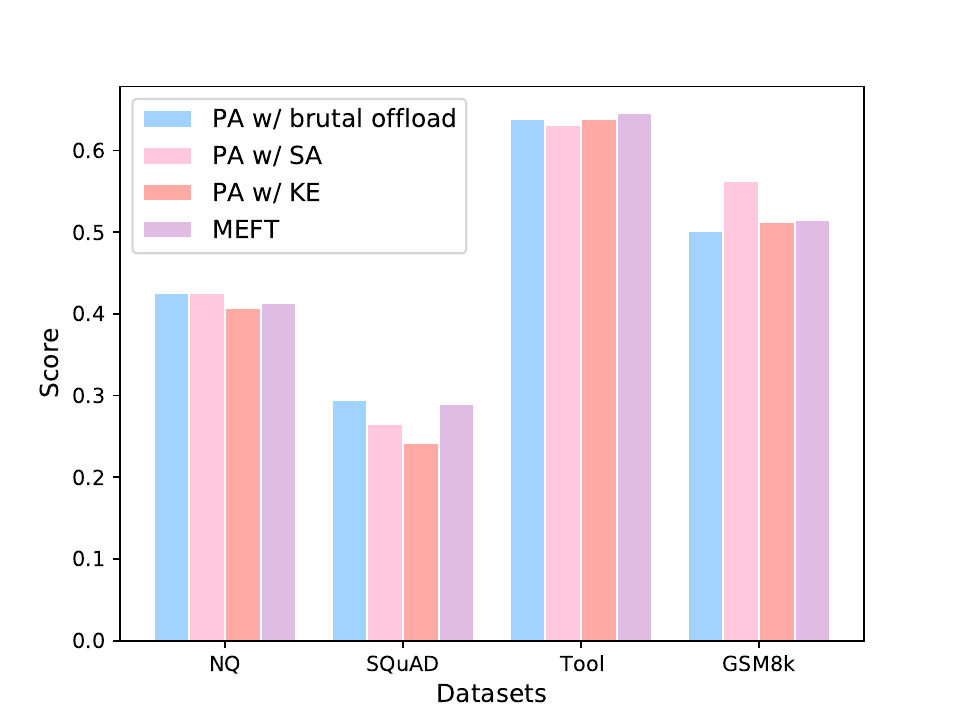}
\caption{Performance comparsion between MEFT w/o KE and MEFT.}
\label{topk_comp_meft}
\end{minipage}\hfill %
\begin{minipage}{.48\textwidth} 
\centering 
\includegraphics[width=\columnwidth]{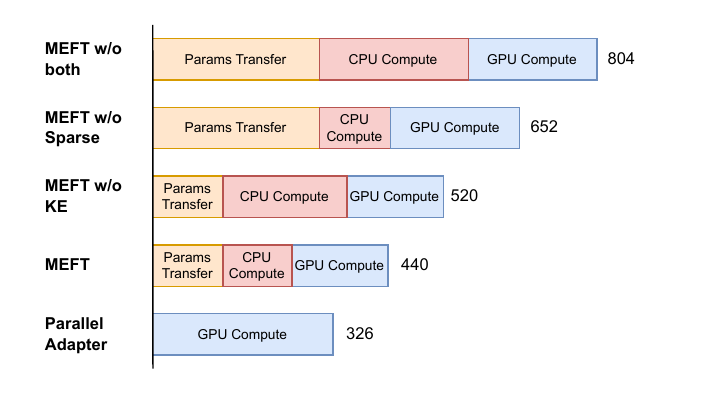}
\caption{Ablation study on latency(ms) per batch relative to Parallel Adapter.}
\label{eff}
\vspace{-4mm}
\end{minipage} 
\end{figure*}
\vskip -0.7in

\section{Experimental Results}

\subsection{Main Results}
Table~\ref{tab:performance_comparsion} presents the main experimental results. Key observations include:
(i) In knowledge-intensive tasks such as NQ, SQuAD, and Tool, our MEFT model significantly outperforms other PEFT approaches when operating under same 24G GPU memory constraints. 
For instance, MEFT achieves 0.413 and 0.427 EM scores on NQ using LLaMA-7B and Mistral-7B models, respectively. 
These scores are notably higher compared to those from baseline methods like Parallel Adapter and LoRA. 
This improvement is attributed to the effective fit of a higher proportion of trainable parameters (i.e., 10\%) within the limited 24G GPU capacity. Moreover, our method also achieves results comparable to full-model fine-tuning.
(ii) When compared with other baselines that also incorporate 10\% trainable parameters, MEFT achieves comparable benchmark performance but only requires 50\% of the GPU memory. This demonstrates MEFT's enhanced efficiency in resource utilization.
(iii) For tasks that are not knowledge-intensive, such as GSM8k, increasing the number of trainable parameters does not yield better performance. However, MEFT's results on GSM8K indicate that sparse training does not compromise performance on these type of task. 
In addition, we conducted experiments where LoRA is attached to FFNs and the detailed results are available in Appendix \ref{sec:lora_ffn}.

\subsection{Ablation Study of Sparse Activation}
It's important to note that the main purpose of using a mechanism similar to MoE (Mixture of Experts) is to alleviate the computational burden on the CPU. The experimental results show that MoE is not the main source of performance improvement.
As shown in Figure~\ref{topk_comp_meft}, partitioned parameters $W_A$ have minimal impact on knowledge-intensive tasks, even improving performance on SQuAD and ToolBench. The term "brutal offload" refers to the straightforward exchange of parameters between the CPU and GPU. However, it exhibits slightly lower performance on GSM8k, a task with a stronger logical component. This suggests that logical tasks may not require a large number of parameters.


\subsection{Efficiency Results}
To analyze the efficiency of our method, Figure~\ref{eff} illustrates the training latency (wall time) per batch, using a server with an RTX 3090 GPU and a 32-core CPU that supports AVX. 
This figure presents the time taken for training across various ablation studies focusing on data transfer, CPU computation, and GPU computation.
Specifically, ``\emph{MEFT w/o both}'' refers to offloading all trainable parameters to CPU and performing computations on the CPU without any optimization, leading to the highest latency observed. 
``\emph{MEFT w/o Sparse}'' offloads too but optimizes by transferring only necessary neurons via PCIe, which cuts down on data transfer time and slightly improves GPU computation efficiency.
``\emph{MEFT w/o KE}'' uses a mixture of experts (MoE) approach for parameter management, reducing computational load but still involves complete parameter transfer.
``\emph{Parallel Adapter}'' performs all operations on the GPU, thus achieving the best latency.

\begin{figure*} 
\centering 
\begin{minipage}{.45\textwidth} 
\centering 
\includegraphics[width=1\linewidth]{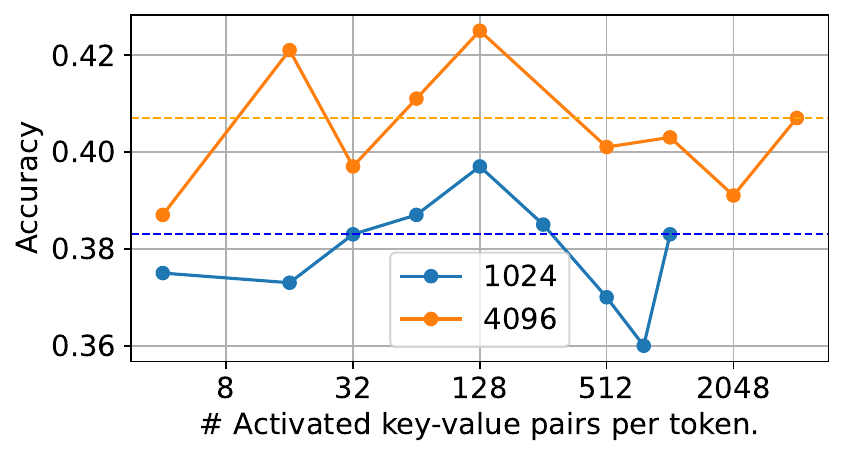}
\caption{Accuracy on NQ respect to the number of activated key-value pairs. Dotted line is the accuracy when $K$= \# key-value pairs.}
\label{topk_progress} 
\end{minipage}\hfill %
\begin{minipage}{.45\textwidth} 
\centering 
\includegraphics[width=1\linewidth]{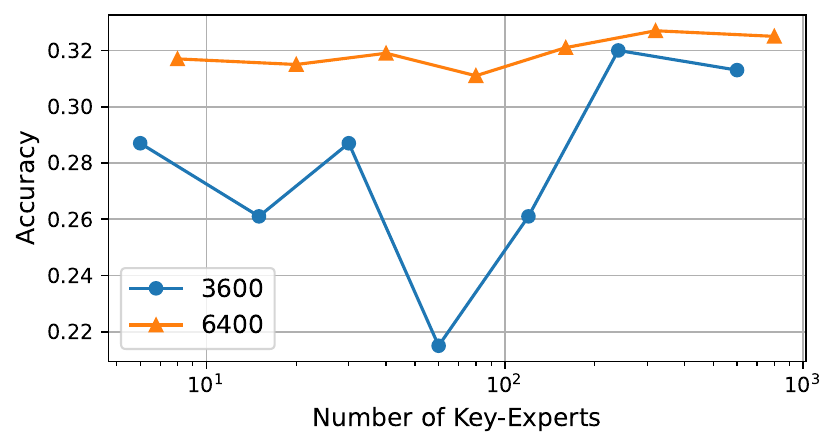}
\caption{Accuracy on Natural Question respect to the number of Key-Experts (Partitions).\newline}
\label{factor_progress}
\vspace{-5mm}
\end{minipage} 
\end{figure*}
\vskip -0.3in

\subsection{Hyperparameters Studies}
Given that numerous parameters may influence performance on downstream tasks, we have tested various hyperparameter settings on the model.

\paragraph{Number of additional key-value pairs} refers to the number of additional parameters that will be trained. The results of LoRA, Adapter, and Parallel Adapter on NQ are shown in Figure~\ref{progress_performace_nq_v1}. It can be observed that when the model needs to be adapted for knowledge-intensive tasks, it often requires more parameters. We also conducted additional tests on the results of Parallel Adapter on SQuAD and ToolBench, as shown in Table~\ref{tab:num_kv_pairs_progress}. These results shows that different datasets require different amounts of additional parameters. SQuAD shows a consistent improvement in performance with an increase in the number of parameters. However, in the NQ results depicted in Figure~\ref{progress_performace_nq_v1}, the best performance was achieved when $3072$ key-value pairs were added to each layer. In contrast, in ToolBench, the best result is achieved when only adding 1024 key-value pairs per layer.

\begin{table}\small
    \centering
    \begin{tabular}{cccc} 
    \toprule
         \textbf{KV pairs} & \textbf{Param.} &  \textbf{SQuAD} & \textbf{ToolBench} \\ 
         \midrule
         128 & 0.4\%&  0.106& 0.604\\ 
         512 & 1.6\%&  0.120& 0.593\\ 
         1024 & 3.2\%&  0.164&  0.636\\ 
         1536 & 4.8\%&  0.236& 0.622\\ 
         2048 & 6.4\%&  0.248& 0.624 \\ 
         4096 & 12.8\%&  0.265& \textbf{0.651}\\ 
         6144 & 19.2\%&  \textbf{0.295} &  0.637\\ 
    \bottomrule
    \end{tabular}
    \caption{The impact of increasing the number of additional key-value pairs per layer to performance.}
    \label{tab:num_kv_pairs_progress}
    \vspace{-4mm}
\end{table}

\paragraph{Number of activated key-value pairs}.
We examined the impact of limiting the number of key-value pairs a token can activate on the final performance of the model. As shown in Figure \ref{topk_progress}, suggest that artificially adding sparsity constraints during training does not substantially affect the performance of the model, when an appropriate value of $K$ is selected. And acceptable performance can be achieved when a single token activates less than 3\% of the parameters.


\paragraph{Number of Key-Experts} refers to the number of partitions in $W_A$. When the number of key experts per layer is equal to the number of additional key-value pairs, each neuron belongs to an independent expert, equivalent to not using MoE partitioning. 
The results, shown in Figure \ref{factor_progress}, align with this theory: the more experts there are (i.e., the more partitions), the better the outcome. At the same time, we find that decent results can be achieved even when the number of experts is small. Consider the case where the number of experts is one, the router $\in \mathbb{R}^d$. The process of retrieving the relevant neuron can be seen as $h$ first dot-multiplying with the router. At this time, all neurons are in expert $E_0$, which is to some extent equivalent to having $r$ partitions.

\subsection{Impact of Batch Size}
\begin{figure}[t]
\begin{center}
\centerline{
\includegraphics[width=1\columnwidth]{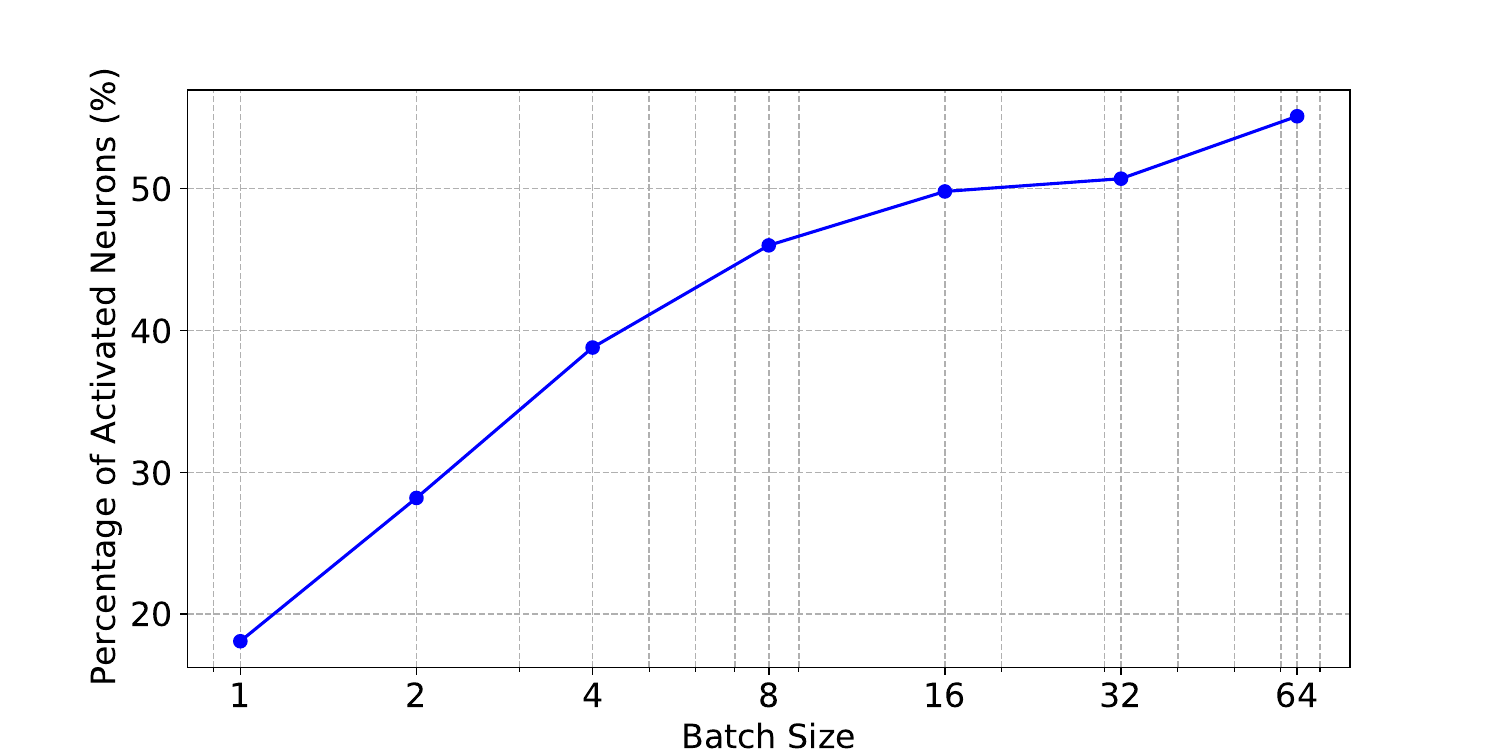}
}
\caption{Percentage of Activated Neurons vs. Batch Size}
\label{fig:bs_percentage}
\end{center}
\vskip -0.3in
\end{figure}

The batch size impacts the weights selected using the top-k method, as the probability of choosing the top-k key-value pairs based on activation values varies among different samples within the batch. Therefore, as the batch size increases, the proportion of activated parameters increases, as shown in Figure \ref{fig:bs_percentage}. It should be noted that our method mainly considers the single-card situation. If a 7B size model is being trained, without using other memory-saving techniques, the maximum batch size that can generally be selected is 4. Therefore, the amount of communication reduced by our method is quite substantial. 

\section{Related Work}
\paragraph{Parameter-Efficient Fine-Tuning (PEFT)} can be categorized into four main types: (1) adapter-based tuning \cite{adapterH, adapterP}, (2) prompt-based tuning \cite{liu2021zhengxiao_prompt1, li2021prefix, schick2020s_prompt3, gao-etal-2021-prompt_n1}, (3) sparse tuning \cite{sung2021training_sparse, vucetic2022efficient_far}, and (4) reparametrization-based tuning \cite{hu2021lora, edalati2022krona_kklora}. 
There are several studies that are relevant to our work. 
\citealp{he2021towards_unified} utilized a unified perspective of previous PEFT methods to propose the Parallel Adapter, which has demonstrated superior performance among most popular PEFT methods \cite{hu2023llm_adapter}. 
In our study, we found the Parallel Adapter to be highly effective in knowledge memorization.
\citealp{wang2022adamix} proposed AdaMix as a method that tunes a mixture of adaptation modules but only with 0.1 - 0.2\% of LLM's parameters.
Furthermore, \citealp{zhang2023lora_fa} preserved the $W_A$ in the LoRA module \cite{hu2021lora} in a frozen state, which resulted in reduced memory usage but didn't try to scale up.

\paragraph{Mixture of Experts} is a computationally efficient architecture that only activates a subset of a large-scale neural network compared to a dense model of the same size \cite{shazeer2017outrageously_moe, fedus2022switch_transformer, dai-etal-2022-stablemoe}. We leverage the computational efficiency of MoE to alleviate the issue of insufficient CPU FLOPs and take advantage of the large memory capacity of the CPU to compensate for its tendency to occupy more memory.

Additional minor related works can be found in Appendix \ref{related_wks}.

\section{Conclusion}
In this paper, we found that the Parallel Adapter can continuously improve performance on knowledge-intensive tasks as parameters increase. We then proposed a memory-efficient training method by leveraging sparsity and MoE. This method significantly reduces the demand for GPU memory and reduces the computational pressure on the CPU. Experimental results show that our method achieves the best results under resource-restricted conditions.

\section*{Limitations}
Although we have tested the impact of the number of added key-value pairs, the number of key-experts, and the size of $\mathbb{K}$ on the model's performance on knowledge-intensive tasks, the effect of this method on the generalization ability of LLMs has not been fully explored. Moreover, this method also lacks testing in the scenario of continuous learning, where it is also suitable. Additionally, the amount of parameters recalled by this method increases with the length of the training sequence, which limits its applicability. We will further explore how to mitigate this phenomenon in our future work.

Due to the absence of detailed memory management and optimization of data transfer between the CPU and GPU (such as custom CUDA streams), the final efficiency of our method is currently not optimal. 

\section*{Ethics Statement}
We acknowledge the importance of the ACM code of Ethics and totally agree with it. We ensure that this work is compatible with the provided code, in terms of the publicly accessed datasets and models.
In particular, this paper focus on the efficient training of Large Language Models (LLMs). Prior research indicates that LLMs can demonstrate biased responses and generate hallucinated information, even when trained on specialized datasets. Therefore, it's crucial to reassess their applicability depend on real-world scenarios.

\section*{Acknowledgements}
This work was supported by the Natural Science Foundation of China (62202271, 61902219, 61972234, 61672324, 62072279, 62102234, 62272274), the National Key R\&D Program of China with grants No. 2020YFB1406704 and No. 2022YFC3303004, the Natural Science Foundation of Shandong Province (ZR2021QF129, ZR2022QF004), the Key Scientific and Technological Innovation Program of Shandong Province (2019JZZY010129), the Fundamental Research Funds of Shandong University, the Tencent WeChat Rhino-Bird Focused Research Program (WXG-FR-2023-07). All content represents the opinion of the authors, which is not necessarily shared or endorsed by their respective employers and/or sponsors.



\bibliography{custom}

\appendix
\begin{figure*}[!t]
\vskip 0.2in
\begin{center}
\centerline{
    \includegraphics[width=\textwidth]{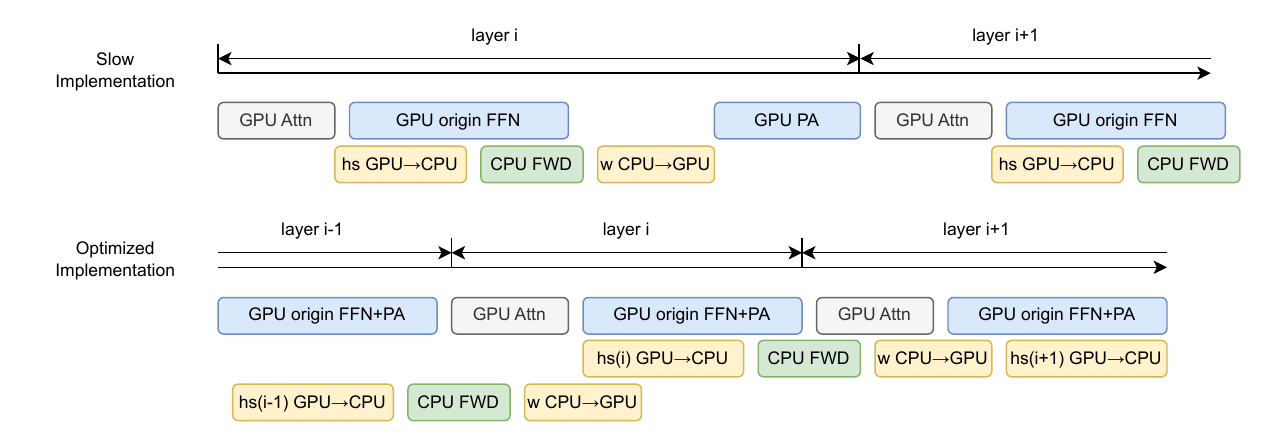}
}
\caption{Slow Implementation and Optimized Implementation}
\label{efficiency_of_pre_retrieval}
\end{center}
\vskip -0.3in
\end{figure*}

\newpage
\section{Implementation Detail and Possible Optimization}
\label{sec:implementation_detail}

\subsection{Implementation Detail}
Figure \ref{efficiency_of_pre_retrieval} shows the general time framework of slow implementation and optimized implementation by Pre-Retrieval. We primarily use Python and PyTorch to implement functions including model forwarding, data transmission, and CPU retrieval of relevant key-value pairs. Specifically, we use Python's multiprocessing API to start multiple processes to read the hidden state transmitted from the GPU, and utilize queues to complete data transmission, achieving asynchronous data computation and transmission, which prevents the GPU from waiting for the CPU to read and calculate.

Common implementations of MoE (Mixture of Experts) such as those in \cite{fedus2022switch_transformer, dai-etal-2022-stablemoe} use a load balance loss to avoid imbalance among experts. In our current implementation, computations for different experts are carried out sequentially on the CPU. Therefore, we have not employed a load balance loss and its impact on performance is unclear at this point.

\subsection{Possible Optimization}
However, there are still areas for optimization in engineering, such as memory management during data transmission, and the relatively low efficiency of Python itself. Additionally, the current model design results in some bubbles in the overall computation process, causing the GPU to wait for the CPU computation to finish.

\section{Negative Attempts}
\label{sec:negative attempts}
The softmax function is commonly used in MoE (Mixture of Experts) to normalize the logits assigned to each expert by the tokens. However, in our scenario, this approach actually led to a decrease in model performance for the NQ and SQuAD tasks during testing.

\section{Other Experiments}
\subsection{LoRA on FFN}
\label{sec:lora_ffn}
Since FFNs are crucial to Transformers, we also attempted to attach LoRA to FFNs. As shown in Table \ref{table:lora_ffn_nq}, LoRA still cannot gradually improve performance with the increase of trainable parameters.

\begin{table}
\centering
\resizebox{\columnwidth}{!}{
    \begin{tabular}{ccccccccc}
    \toprule
    Rank & 4 & 8 & 16 & 64 & 128 & 256 & 512 & 1024 \\
    \hline
    Acc & 0.297 & 0.293 & 0.279 & 0.289 & 0.281 & 0.277 & 0.281 & 0.293 \\
    \bottomrule
    \end{tabular}
}
\caption{Performance of LoRA-FFN on NQ.}
\vskip -0.2in
\label{table:lora_ffn_nq}
\end{table}

\section{Potential Risks}
LLMs' stronger memory of knowledge could potentially lead to a stronger memory of privacy or other sensitive information. With the ongoing research into language model attacks \cite{birch2023model_attack,niu2024jailbreaking_attack}, this enhanced memory of knowledge by LLMs could result in greater leakage of private information, particularly if trained on private datasets.

\section{Other Related Works}
\label{related_wks}
\paragraph{Other PEFTs}. 
Additionally, \citealp{diao2023mixture} attempted to incorporate domain-specific knowledge by adopting a mixture-of-adapters mechanism, but only 0.7\% of the trainable parameters of RoBERTa-large \cite{liu2019roberta} were utilized, limiting the model's capacity.

\paragraph{Sparsity of LLM} has been partially explored with the objective of accelerating inference. SparseGPT \cite{frantar2023sparsegpt} demonstrated that LLMs can be pruned to a sparsity level of at least 50\% with only a minimal loss in accuracy. The work of \citet{mirzadeh2023relu_strikes} indicated that the ReLU activation function, through sparse activation, significantly reduces computation and weight transfer. Deja Vu \cite{liu2023dejavu} identified the presence of contextual sparsity for any given input and illustrated that a sparsity level of at least 95\% can be achieved within the FFN Block. We apply these findings on sparsity to the fine-tuning process.

\paragraph{Memory Efficient Fine-tune Methods}. The main feature of QLoRA \cite{dettmers2024qlora} is to reduce memory usage by quantizing the model, and quantization is orthogonal to our method, so we can further quantize the parameters exchanged between the CPU and GPU.
\citealp{liao2024make_meft} primarily reduces the memory usage of intermediate activations by reversing the model. However, in the scenarios we consider, when more trainable parameters are added, the memory consumed by these parameters and their corresponding optimizer states already exceeds the total.
LST \cite{lst2022} constructs a "bypass" based on the original large model, making it unnecessary to perform backpropagation directly on the original large model. 
Deepspeed's Zero \cite{ren2021zero_offload_deepspeed}  uses offload technology to exchange parameters between the CPU and GPU, which is similar to our method. However, our method utilizes sparsity, selecting relevant parameters based on context to reduce the communication pressure between the CPU and GPU, and uses the Mixture of Experts (MoE) to reduce the computational pressure on the CPU.

\paragraph{Neural Memory} proposed by \citet{sukhbaatar2015end_neural_memory} consists of $n$ key-value pairs, with both key and value represented by a $d$-dimensional vector $k_i, v_i \in \mathbb{R}^{d}$, forming $K, V \in \mathbb{R}^{n \times d}$. Then $MN(x)=\text{softmax}(x\cdot K^T)\cdot V$ is used to acquire knowledge related to $x$. \citet{geva-etal-2021-transformer} proposed that the FFN Block in Transformers can emulate Neural Memory, suggesting that $k_i$ captures the pattern of the input sequence, while the value represents the distribution of the token. \citet{knowledge_neur} demonstrated that factual knowledge is stored in what they term 'knowledge neurons', and that editing these neurons can modify the stored knowledge. Similarly, ROME \cite{meng2022locating_rome} found that knowledge is stored in the middle-layer feed-forward modules. They used causal intervention to identify decisive neuron activations. We leveraged the key-value format of knowledge neurons in conjunction with the sparsity of LLMs, which significantly reduced the memory overhead during the fine-tuning phase.

\paragraph{Approximate Nearest Neighbor Search} algorithms sacrifice recall to improve the efficiency of nearest neighbor searches. However, popular methods such as HNSW \cite{hnsw}, Faiss~\cite{johnson2019billion_faiss}, and ScaNN \cite{avq_2020_scann} have a low tolerance for frequently updated vectors, as this affects the pre-built retrieval structure. Since backpropagation will result in the update of all activated FFN neurons, we did not use ANN. Although some studies \cite{chen2020slide, chen2021mongoose, zeng2023lookupffn} have achieved comparable or even faster CPU training speeds than the GPU by applying LSH, these methods are applicable to rather specific network configurations. For instance, the method in \citealp{chen2020slide} is only suitable when the network is very wide. Our method, however, significantly reduces the overall computation volume by partitioning the parameter matrix in a way similar to MoE. We will delve into the specifics in the following analysis.
However, popular methods such as HNSW \cite{hnsw}, Faiss \cite{johnson2019billion_faiss}, and ScaNN \cite{avq_2020_scann} have a low tolerance for frequently updated vectors, as this affects the pre-built retrieval structure. Since backpropagation will result in the update of all activated FFN neurons, we propose a Key-Experts (KE) mechanism to tackle this issue.

\end{document}